%% file: COLI-manual3.tex
\documentclass{clv3}

\usepackage{hyperref}
\usepackage{xcolor}
\usepackage{wrapfig}
\definecolor{darkblue}{rgb}{0, 0, 0.5}
\hypersetup{colorlinks=true,citecolor=darkblue, linkcolor=darkblue, urlcolor=darkblue}

\usepackage{booktabs}
\usepackage{amsmath}
\usepackage[utf8]{inputenc}
\usepackage{tikz}
\usetikzlibrary{arrows}
\usepackage{amssymb}
\usetikzlibrary{arrows.meta}

\colorlet{MyColorOne}{blue!50}
\colorlet{MyColorTwo}{red!50}

\newcommand{\lightercolor}[3]{
    \colorlet{#3}{#1!#2!white}
}

\lightercolor{MyColorOne}{50}{BlueLight}
\lightercolor{MyColorTwo}{50}{PinkLight}

\issue{1}{1}{2018}

\runningtitle{What do Language Representations Really Represent?}

\runningauthor{Bjerva et al.}

\begin{document}

\title{What do Language Representations Really Represent?}

\author{Johannes Bjerva\thanks{E-mail: bjerva@di.ku.dk}}
\affil{Department of Computer Science, University of Copenhagen}

\author{Robert Östling}
\affil{Department of Linguistics,\\Stockholm University}

\author{Maria Han Veiga}
\affil{Institute of Computational Science, University of Zurich}

\author{Jörg Tiedemann}
\affil{Department of Digital Humanities, University of Helsinki}

\author{Isabelle Augenstein}
\affil{Department of Computer Science, University of Copenhagen}

\maketitle

\begin{abstract}
A neural language model trained on a text corpus can be used to induce distributed representations of words, such that similar words end up with similar representations. If the corpus is multilingual, the same model can be used to learn distributed representations of languages, such that similar languages end up with similar representations. We show that this holds even when the multilingual corpus has been translated into English, by picking up the faint signal left by the source languages.
However, just like it is a thorny problem to separate semantic from syntactic similarity in word representations, it is not obvious what type of similarity is captured by language representations. We investigate correlations and causal relationships between language representations learned from translations on one hand, and genetic, geographical, and several levels of structural similarity between languages on the other. Of these, structural similarity is found to correlate most strongly with language representation similarity, while genetic relationships---a convenient benchmark used for evaluation in previous work---appears to be a confounding factor. Apart from implications about translation effects, we see this more generally as a case where NLP and linguistic typology can interact and benefit one another.
\end{abstract}

\section{Introduction}

\begin{wrapfigure}{r}{0.36\textwidth}
  \begin{center}
    \includegraphics[width=0.35\textwidth]{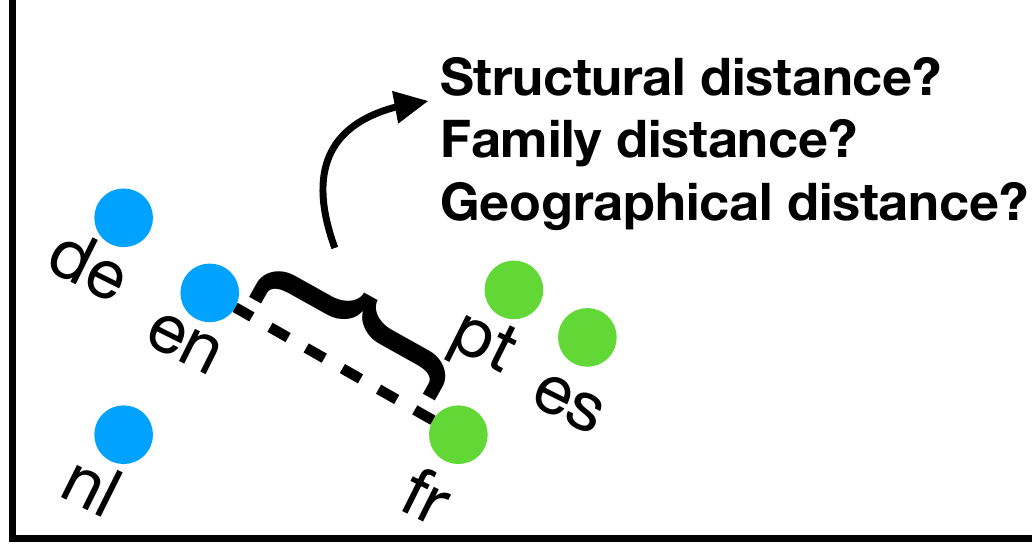}
  \end{center}
    \caption{\label{fig:similarities}Language representations in a two-dimensional space. What do their similarities represent?}
\end{wrapfigure}

Words can be represented with distributed word representations, currently often in the form of word embeddings. 
Similarly to how words can be embedded, so can languages, by associating each language with a real-valued vector known as a \textit{language representation}, which can be used to measure similarities between languages. 
This type of representation can be obtained by, e.g., training a multilingual model for some NLP task \citep{ostling_tiedemann:2017,malaviya:2017,googlenmt}.
The focus of this work is on the evaluation of similarities between such representations.
This is an important area of work, as computational approaches to typology \citep{dunn:2011,cotterell:2017,bjerva:2018:naacl} have the potential to answer research questions on a much larger scale than traditional typological research \citep{haspelmath:2001}.
Furthermore, having knowledge about the relationships between languages can help in NLP applications \citep{ammar:2016}, and having incorrect interpretations can be detrimental to multilingual NLP efforts.
For instance, if the similarities between languages in an embedded language space were to be found to encode geographical distances (Figure~\ref{fig:similarities}), any conclusions drawn from use of these representations would not likely be of much use for most NLP tasks.
The importance of having deeper knowledge of what such representations encapsulate is further hinted at by both experiments with interpolation of language vectors in \citet{ostling_tiedemann:2017}, as well as multilingual translation models \citep{googlenmt}.

Several previous authors have done preliminary investigations into the structure of language representations: \citet{ostling_tiedemann:2017}, \citet{malaviya:2017} and \citet{googlenmt} in the context of language modelling and machine translation, all of them using multilingual data. In this work we follow up on the findings of \citet{haifa_transl:2017} who, by using language representations consisting of manually specified feature vectors, find that the structure of a language representation space is approximately preserved by translation. However, their analysis only stretches as far as finding a correlation between their language representations and genetic distance, even though the latter is correlated to several other factors.
We apply a multilingual language model to this problem, and evaluate the learned representations against a set of three language properties: (i) genetic distance (families), (ii) a novel measure of syntactic similarity (structural), and (iii) distance of language communities (geographical).
We investigate:
    \vspace{-0.35cm}
    \paragraph{RQ1} In which way do different language representations encode language similarities? In particular, is genetic similarity what is really captured?
    \vspace{-0.35cm} 
    \paragraph{RQ2} What causal relations can we find between language representation similarities?
    \vspace{-0.2cm}

\subsection{Contributions}
Our work is most closely related to \citet{haifa_transl:2017} who investigate representation learning on monolingual English sentences, which are translations from various source languages to English from the Europarl corpus \citep{europarl}.
They employ a feature-engineering approach to predict source languages and learn an Indo-European (IE) family tree using their language representations, showing that there are significant traces of the source languages in translations.
They use features based on sequences of POS tags, function words and cohesive markers.
Additionally they posit that the similarities found between their representations encode the genetic relationships between languages.
We show that this is not the strongest explanation of the similarities as a novel syntactic measure offers far more explanatory value, which we further substantiate by investigating causal relationships between language representations and similarities \citep{causality}.
This is an important finding as it highlights the need for thoroughly substantiating linguistic claims made based on empirical findings. Further, understanding what similarities are encoded in language embeddings gives insights into how language embeddings could be used for downstream multilingual NLP tasks. If language representations are used for transfer learning to low-resource languages, having an incorrect view of the structure of the language representation space can be dangerous. For instance, the standard assumption of genetic similarity would imply that the representation of the Gagauz language (Turkic, spoken mainly in Moldova) should be interpolated from the genetically very close Turkish, but this would likely lead to poor performance in syntactic tasks since the two languages have diverged radically in syntax relatively recently.

\begin{figure*}[tb]
	\centering
    \includegraphics[width=0.91\textwidth]{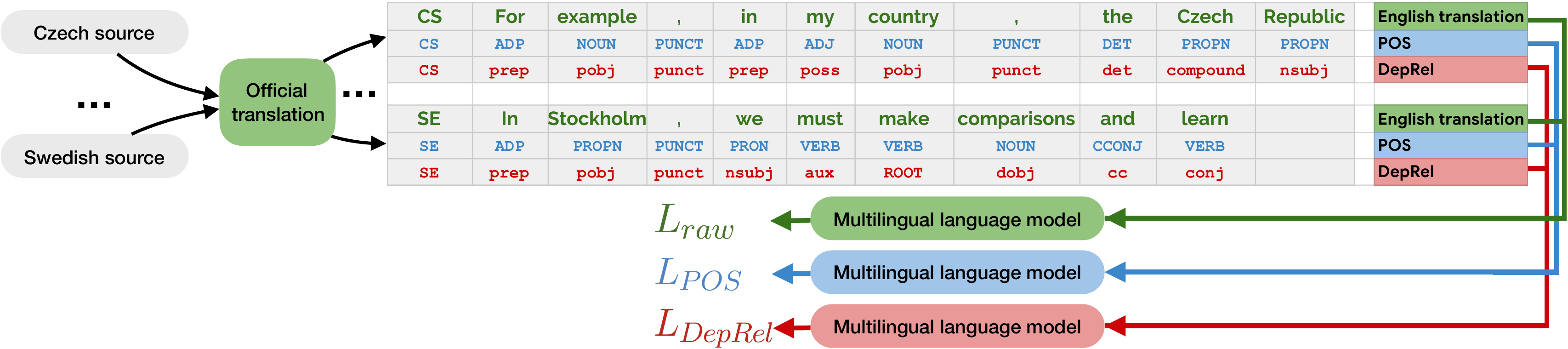}
    \caption{\label{fig:problem}Problem illustration. Given official translations from EU languages to English, we train multilingual language models on various levels of abstractions, encoding the source languages. The resulting source language representations ($L_{raw}$ etc.) are evaluated.}
\end{figure*}

\section{Method}

Figure~\ref{fig:problem} illustrates the data and problem we consider in this paper.
We are given a set of English gold-standard translations from the official languages of the European Union, based on speeches from the European Parliament.\footnote{This is the exact same data as used by \citet{haifa_transl:2017}, originating from Europarl \citep{europarl}.}
We wish to learn language representations based on this data, and investigate the linguistic relationships which hold between the resulting representations (\textbf{RQ1}).
It is important to abstract away from the surface forms of the translations as, e.g., speakers from certain regions will tend to talk about the same issues, or places.
We therefore introduce three levels of abstraction:
i) training on function words and POS;
ii) training on only POS tags (POS in Figure~\ref{fig:problem});
iii) training on sequences of dependency relation tags (DepRel in Figure~\ref{fig:problem}), and constituent tags.
This annotation is obtained using UDPipe \citep{udpipe}.

\subsection{Language Representations}
For each level of abstraction, we train a multilingual neural language model, in order to obtain representations (vectors in $\mathbb{R}^k$) which we can analyse further (\textbf{RQ1}).
Note that this model is multilingual in the sense that we model the \textit{source language} of each input sequence, whereas the input sequences themselves are, e.g., sequences of POS tags.
Our model is a multilingual language model using a standard 2-layer LSTM architecture.
Multilinguality is approached similarly to \citet{ostling_tiedemann:2017} who include a language representation at each time-step.
That is to say, each input is represented both by a symbol representation, $c$, and a language representation, $l\in L$.
Since the set of language representations $L$ is updated during training, the resulting representations encode linguistic properties of the languages.
Whereas \citet{ostling_tiedemann:2017} model hundreds of languages, we model only English - however, we redefine $L$ to be the set of source languages from which our translations originate.

\section{Family Trees from Translations}
We now consider the language representations obtained from training our neural language model on the input sequences with different representations of the text (characters, POS sequences, etc.).
We cluster the language representations---vectors in $\mathbb{R}^k$---hierarchically\footnote{Following \citet{haifa_transl:2017} we use the same implementation of Ward's algorithm. We use vector cosine distance rather than Euclidean distance because it is more natural for language vector representations where the vector magnitude is not important.} 
and compute similarities between our generated trees and the gold tree of \citet{serva:2008}, using the distance metric from \citet{haifa_transl:2017}.\footnote{Trees not depicted here can be found in the supplements.}
Our generated trees yield comparable results to previous work (Table~\ref{tab:tree_eval}).

\begin{wraptable}{r}{0.58\textwidth}
\small
  \begin{tabular}{lrr}
  	\toprule
  	\textbf{Condition}  & \textbf{Mean} & \textbf{St.d.} \\
    \midrule
    Raw text (LM-Raw)                 & 0.527          & -   \\
    Function words and POS (LM-Func)                & 0.556          & -   \\
    Only POS (LM-POS)                 & 0.517          & -   \\
    Phrase-structure (LM-Phrase)              & 0.361          & -   \\
    Dependency Relations (LM-Deprel)              & \textbf{0.321} & -   \\
    \midrule
    \textit{POS trigrams} (ROW17) & 0.353          & 0.06 \\
    \textit{Random}	(ROW17)	   & 0.724          & 0.07 \\
  \end{tabular}   \caption{\label{tab:tree_eval}Tree distance evaluation (lower is better, cf. §5.1). } 
\end{wraptable}

\paragraph{Language modelling using lexical information and POS tags}

Our first experiments deal with training directly on the raw translated texts.
This is likely to bias representations by speakers from different countries talking about specific issues or places (as in Figure~\ref{fig:problem}), and gives the model comparatively little information to work with as there is no explicit syntactic information available.
As a consequence of the lack of explicit syntactic information, it is unsurprising that the results (\textbf{LM-Raw} in Table~\ref{tab:tree_eval}) only marginally outperform the random baseline.

To abstract away from the content and negate the geographical effect we train a new model on only function words and POS.
This performs almost on par with LM-Raw (\textbf{LM-Func} in Table~\ref{tab:tree_eval}), indicating that the level of abstraction reached is not sufficient to capture similarities between languages. 
We next investigate whether we can successfully abstract away from the content by removing function words, and only using POS tags (\textbf{LM-POS} in Table~\ref{tab:tree_eval}).
Although \citet{haifa_transl:2017} produce sensible trees by using trigrams of POS and function words, we do not obtain such trees in our most similar settings.
One hypothesis for why this is the case, is the differing architectures used, indicating that our neural architecture does not pick up on the trigram-level statistics present in their explicit feature representations.

\paragraph{Language Modelling on phrase structure trees and dependency relations}

To force the language model to predict as much syntactic information as possible, we train on bracketed phrase structure trees.
Note that this is similar to the target side of \citet{Vinyals2015GFL}.
All content words are replaced by POS tags, while function words are kept.
This results in a vocabulary of 289 items (phrase and POS tags and function words).
Syntactic information captures more relevant information for reconstructing trees than previous settings (\textbf{LM-Phrase} in Table~\ref{tab:tree_eval}), yielding trees of similar quality to previous work.

\begin{figure*}[t]
\centering
    \includegraphics[width=0.92\textwidth]{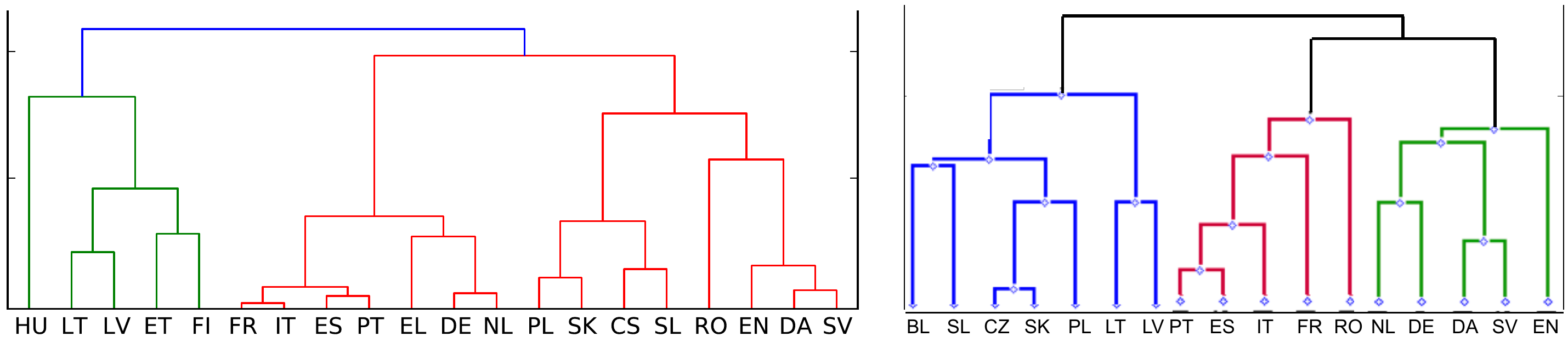}
    \caption{\label{fig:ud_clustering}Clustering based on dependency link statistics from UD (left), and the genetic tree from \citet{serva:2008} (right). Which type of similarity do language representations really represent?}
\end{figure*}

We also compare to the UD dependency formalism, as we train the language model on tuples encoding the dependency relation and POS tag of a word, the head direction, and the head POS tag (\textbf{LM-Deprel} in Table~\ref{tab:tree_eval}).
The \textbf{LM-Phrase} and \textbf{LM-Deprel} models yield the best results overall, due to them having access to higher levels of abstraction via syntax.
The fact that sufficient cues for the source languages can be found here shows that source language affects the grammatical constructions used (cf.~\citet{translationese}). 

\section{Comparing Languages}

\label{sec:comparinglanguages}
Our main contribution is to investigate whether genetic distance between languages which is captured by language representations, or if other distance measures provide more explanation (\textbf{RQ1}).
Having shown that our language representations can reproduce genetic trees on-par with previous work, we now compare the language embeddings using three different types of language distance measures: \textit{genetic distance} estimated by methods from historical linguistics, \textit{geographical distance} of speaker communities, and a novel measure for the \textit{structural distances} between languages.

\subsection{Genetic Distance}
Following \citet{haifa_transl:2017}, we use phylogenetic trees from \citet{serva:2008} as our gold-standard representation of genetic distance (Figure~\ref{fig:ud_clustering}).
For meaningful and fair comparison, we also use the same distance metric.
The metric considers a tree of $N$ leaves, $l_n$.
The weighted distance between two leaves in a tree $\tau$, denoted $D_\tau(l_n, l_m)$, is the sum of the weights of all edges on the shortest path between these leaves.
The distance between a generated tree, $g$, and the gold tree, $\tau$, can then be calculated by summing the square of the differences between all leaf-pair distances \citep{haifa_transl:2017}:
\begin{equation*}
  Dist(\tau, g) = \sum_{n,m\in N} (D_\tau(l_n,l_m)-D_g(l_n,l_m))^2.
\end{equation*}

\subsection{Geographical Distance}
We rely on the coordinates provided by Glottolog \citep{glottolog}. 
These are by necessity approximate, since the geography of a language cannot accurately be reduced to a single point denoting the geographical centre point of where its speakers live.
Still, this provides a way of testing the influence of geographical factors such as language contact or political factors affecting the education system.

\subsection{Structural Distance}

To summarise the structural properties of each language, we use counts of dependency links from the Universal Dependencies treebanks (UD), version 2.1 \citep{ud21}. Specifically, we represent each link by combining head and dependent POS, dependency type, and direction.
This yields 8607 combinations, so we represent each language by a 8607-dimensional normalised vector, and compute the cosine distance between these language representations.

Figure~\ref{fig:ud_clustering} shows the result of clustering these vectors (Ward clustering, cosine distance).
While strongly correlated with genealogical distance, significant differences can be observed. Romanian, as a member of the Balkan sprachbund, is distinct from the other Romance languages. The North Germanic (Danish, Swedish) and West Germanic (Dutch, German) branches are separated due to considerable structural differences, with English grouped with the North Germanic languages despite its West Germanic origin. The Baltic languages (Latvian, Lithuanian) are grouped with the nearby Finnic languages (Estonian, Finnish) rather than their distant Slavic relatives.

This idea has been explored previously by \citet{conf/depling/ChenG17}, who use a combination of relative frequency, length and direction of deprels. 
We, by comparison, achieve an even richer representation by also taking head and dependent POS into account. 

\section{Analysis of Similarities}

\begin{wrapfigure}{r}{0.5\textwidth}
\vspace{-20pt}
  \begin{center}
    \includegraphics[width=0.43\textwidth]{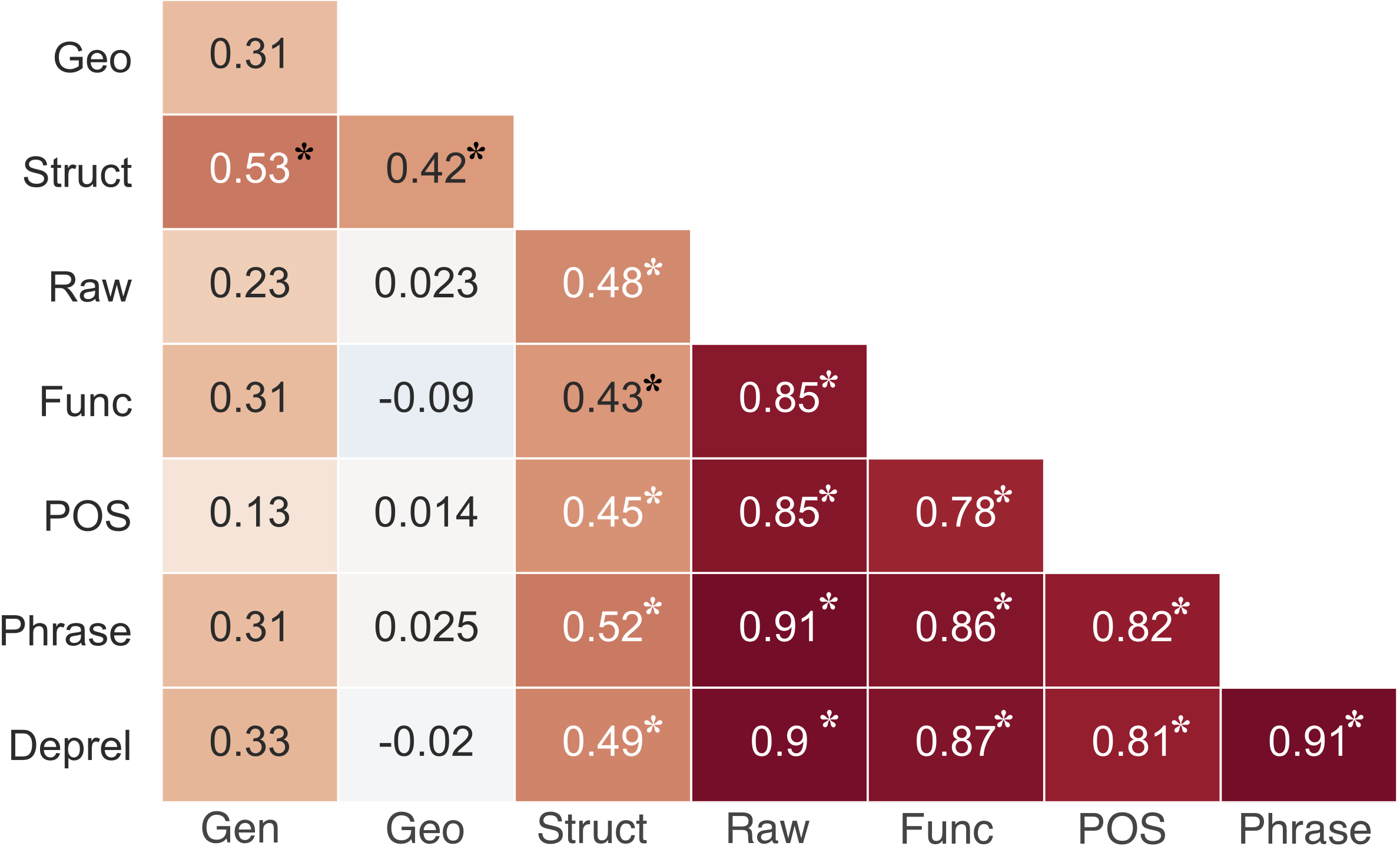}
  \end{center}
  \caption{\label{fig:correlations}Correlations between similarities (Genetic, Geo., and Struct.) and language representations (Raw, Func, POS, Phrase, Deprel).
  Significance at $p<0.001$ is indicated by *.}
\vspace{-5pt}
\end{wrapfigure}

Although we are able to reconstruct phylogenetic language trees in a similar manner to previous work, we wish to investigate whether genetic relationships between languages really is what our language representations represent.

We generate distance matrices $A_{\rho}$, where each entry $a_{i,j}$ represents the $\rho$-similarity between the $i^{th}$ and $j^{th}$ languages, using the three similarity measures outlined in ~§4. Then, the entries in $A_{gen}$ contain pairwise genetic distances, computed by summing the weights of all edges on the shortest path between two leaves (languages). Similarly, the entries in $A_{geo}$ contain the geographical distance between countries associated with the languages. Lastly, the entries in $A_{struct}$ contain the cosine distance between the language representations, which are encoded in 8607-dimensional normalised vectors.

Figure~\ref{fig:correlations} shows the Spearman correlation coefficients between each pair of these matrices. The strongest correlations can be found between the language embeddings, showing that they have similar representations.
The correlations between our three distance measures are also considerable, e.g., between geographical and structural distances.
This is expected, as languages which are close to one another geographically tend to be similar due to language contact, and potentially shared origins \citep{velupillai:2012}.

\paragraph{What do language representations really represent?\hspace{-0.4em}}
Most interestingly, the language embedding similarities correlate the most strongly with the structural similarities, rather than the genetic similarities, thus answering \textbf{RQ1}.
Although previous work by \citet{haifa_transl:2017} has shown that relatively faithful phylogenetic trees can be reconstructed, we have found an alternative interpretation to these results with much stronger similarities to structural similarities.
This indicates that, as often is the case, although similarities between two factors can be found, this is not necessarily the factor with the highest explanatory value \citep{roberts:2013}.

\section{Causal Inference}
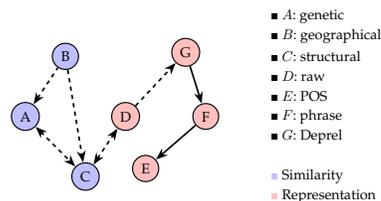
\begin{wrapfigure}{r}{0.4\textwidth}
  \resizebox{0.38\textwidth}{!}{
  	\input{figure}
  }
  \caption{\label{fig:network}Causal network generated by IC*.}
\vspace{-10pt} 
\end{wrapfigure}

We further strengthen our analysis by investigating \textbf{RQ2}, looking at the relationships between our variables in a Causal Network \citep{causality}.
We use a variant of the Inductive Causation algorithm, namely IC* \citep{verma:1992}.
It takes a distribution as input, and outputs a partially directed graph which denotes the (potentially) causal relationships found between each node in the graph.
Here, the nodes represent our similarity measures and language embedding distances.
The edges in the resulting graph can denote genuine causation (unidirectional edges), potential causation (dashed unidirectional edges), spurious associations (bidirectional edges), and undetermined relationships (undirected edges) \citep{causality}.
Running the algorithm on our distribution based on all the distance measures and language embeddings from this work yields a graph with the following properties, as visualised in Figure~\ref{fig:network}.\footnote{The IC* algorithm uses pairwise correlations to find sets of conditional independencies between variables at $p<0.001$, and constructs a minimal partially directed graph which is consistent with the data.}
We observe two clusters, marking associations between distance measures, and language representations.
Interestingly, the only link found between the clusters is an association between the structural similarities and our raw model.
This further strengthens our argument, as the fact that no link is found to the genetic similarities shows that our alternative explanation has higher explanatory value, and highlights the need for controlling for more than a single linguistic factor when seeking explanations for ones results.

\section{Discussion and Conclusions}
We train language representations on three levels of syntactic abstraction, and explore three different explanations to what language representations represent: genetic, geographical, and structural distances.
On the one hand, we extend on previous work by showing that phylogenetic trees can be reconstructed using a variety of language representations \citep{haifa_transl:2017}.
On the other, contrary to a claim of \citet{haifa_transl:2017}, we show that structural similarities between languages are a better predictor of language representation similarities than genetic similarities.
As interest in computational typology is increasing in the NLP community \citep{ostling:2015,bjerva:2018:naacl,ctsurvey,gerzrelation}, we advocate for the necessity of explaining typological findings through comparison.

\starttwocolumn
\bibliography{naaclhlt2018}

\end{document}

%% file: figure.tex
\begin{tikzpicture}
\begin{scope}[every node/.style={circle,thick,draw}, NodeS/.style={circle,fill=BlueLight}, NodeLR/.style={circle,fill=PinkLight}]
    \node[NodeS] (A) at (0,1) {A};
    \node[NodeS] (B) at (1,2.5) {B};
    \node[NodeS] (C) at (1.5,-0.5) {C};
    \node[NodeLR] (D) at (2.5,1) {D};
    \node[NodeLR] (F) at (3,-0.3) {E} ;
    \node[NodeLR] (G) at (4.5,1) {F} ;
    \node[NodeLR] (H) at (4,2.6) {G} ;
\end{scope}

\begin{scope}[>={Stealth[black]},
              every edge/.style={draw=black,very thick}, edgeWeak/.style={draw=black,dashed},
              edgeStrong/.style={draw=black}]
    \path[edgeWeak] [<-] (A) edge node {} (B); 
    \path[edgeWeak] [<->] (A) edge node {} (C); 
    \path[edgeWeak] [->] (B) edge node {} (C); 
    \path[edgeWeak] [<->] (C) edge node {} (D); 
    \path[edgeWeak] [->] (D) edge node {} (H); 
    \path[edgeStrong] [->] (G) edge node {} (F); 
    \path[edgeStrong] [->] (H) edge node {} (G); 
\end{scope}
\begin{scope}[shift={(6cm,4cm)}]
        \node[anchor=west] at (0,-0.5) {{\tiny$\blacksquare$} $A$: genetic};
        \node[anchor=west] at (0,-1) {{\tiny$\blacksquare$} $B$: geographical};
        \node[anchor=west] at (0,-1.5) {{\tiny$\blacksquare$} $C$: structural};
        \node[anchor=west] at (0,-2) {{\tiny$\blacksquare$} $D$: raw};        
        \node[anchor=west] at (0,-2.5) {{\tiny$\blacksquare$} $E$: POS};        \node[anchor=west] at (0,-3) {{\tiny$\blacksquare$} $F$: phrase};
        \node[anchor=west] at (0,-3.5) {{\tiny$\blacksquare$} $G$: Deprel};
        \node[anchor=west] at (0,-4.5) {{\textcolor{BlueLight}{\tiny$\blacksquare$}} Similarity};
\node[anchor=west] at (0,-5) {{\textcolor{PinkLight}{\tiny$\blacksquare$}} Representation};
\end{scope}
\end{tikzpicture}

%% file: COLI-manual3.bbl
\begin{thebibliography}{24}
\expandafter\ifx\csname natexlab\endcsname\relax\def\natexlab#1{#1}\fi

\bibitem[{Ammar et~al.(2016)Ammar, Mulcaire, Ballesteros, Dyer, and
  Smith}]{ammar:2016}
Ammar, Waleed, George Mulcaire, Miguel Ballesteros, Chris Dyer, and Noah Smith.
  2016.
\newblock Many languages, one parser.
\newblock \emph{TACL}, 4:431--444.

\bibitem[{Bjerva and Augenstein(2018)}]{bjerva:2018:naacl}
Bjerva, Johannes and Isabelle Augenstein. 2018.
\newblock From phonology to syntax: Unsupervised linguistic typology at
  different levels with language embeddings.
\newblock In \emph{NAACL-HLT}.

\bibitem[{Chen and Gerdes(2017)}]{conf/depling/ChenG17}
Chen, Xinying and Kim Gerdes. 2017.
\newblock {Classifying Languages by Dependency Structure. Typologies of
  Delexicalized Universal Dependency Treebanks}.
\newblock In \emph{DepLing}, pages 54--63.

\bibitem[{Cotterell and Eisner(2017)}]{cotterell:2017}
Cotterell, Ryan and Jason Eisner. 2017.
\newblock {Probabilistic Typology: Deep Generative Models of Vowel
  Inventories}.
\newblock In \emph{ACL}.

\bibitem[{Dunn et~al.(2011)Dunn, Greenhill, Levinson, and Gray}]{dunn:2011}
Dunn, Michael, Simon~J Greenhill, Stephen~C Levinson, and Russell~D Gray. 2011.
\newblock Evolved structure of language shows lineage-specific trends in
  word-order universals.
\newblock \emph{Nature}, 473(7345):79--82.

\bibitem[{Gellerstam(1986)}]{translationese}
Gellerstam, Martin. 1986.
\newblock Translationese in swedish novels translated from english.
\newblock \emph{Translation studies in Scandinavia}, 1:88--95.

\bibitem[{Gerz et~al.(2018)Gerz, Vulic, Ponti, Reichart, and
  Korhonen}]{gerzrelation}
Gerz, Daniela, Ivan Vulic, Edoardo~Maria Ponti, Roi Reichart, and Anna
  Korhonen. 2018.
\newblock {On the Relation between Linguistic Typology and (Limitations of)
  Multilingual Language Modeling}.
\newblock In \emph{{EMNLP}}.

\bibitem[{Hammarstr{\"o}m, Forkel, and Haspelmath(2017)}]{glottolog}
Hammarstr{\"o}m, Harald, Robert Forkel, and Martin Haspelmath. 2017.
\newblock Glottolog 3.0.
\newblock \emph{Jena: Max Planck Institute for the Science of Human History.
  (Available online at http://glottolog.org, accessed on 2017-05-15.)}.

\bibitem[{Haspelmath(2001)}]{haspelmath:2001}
Haspelmath, Martin. 2001.
\newblock \emph{{Language typology and language universals: An international
  handbook}}, volume~20.
\newblock Walter de Gruyter.

\bibitem[{Johnson et~al.(2017)Johnson, Schuster, Le, Krikun, Wu, Chen, Thorat,
  Vi{\'e}gas, Wattenberg, Corrado, Hughes, and Dean}]{googlenmt}
Johnson, Melvin, Mike Schuster, Quoc~V. Le, Maxim Krikun, Yonghui Wu, Zhifeng
  Chen, Nikhil Thorat, Fernanda Vi{\'e}gas, Martin Wattenberg, Greg Corrado,
  Macduff Hughes, and Jeffrey Dean. 2017.
\newblock {Google's Multilingual Neural Machine Translation System: Enabling
  Zero-Shot Translation}.
\newblock \emph{TACL}, 5:339--351.

\bibitem[{Koehn(2005)}]{europarl}
Koehn, Philipp. 2005.
\newblock {Europarl: A Parallel Corpus for Statistical Machine Translation}.
\newblock In \emph{{MT Summit X}}.

\bibitem[{Malaviya, Neubig, and Littell(2017)}]{malaviya:2017}
Malaviya, Chaitanya, Graham Neubig, and Patrick Littell. 2017.
\newblock {Learning Language Representations for Typology Prediction}.
\newblock In \emph{EMNLP}, pages 2519--2525.

\bibitem[{Nivre et~al.(2017)}]{ud21}
Nivre, Joakim et~al. 2017.
\newblock {Universal Dependencies 2.1}.
\newblock {LINDAT}/{CLARIN} digital library at the Institute of Formal and
  Applied Linguistics ({{\'U}FAL}), Charles University.

\bibitem[{{\"O}stling(2015)}]{ostling:2015}
{\"O}stling, Robert. 2015.
\newblock {Word Order Typology through Multilingual Word Alignment}.
\newblock In \emph{ACL-IJCNLP}, pages 205--211.

\bibitem[{{\"O}stling and Tiedemann(2017)}]{ostling_tiedemann:2017}
{\"O}stling, Robert and J{\"o}rg Tiedemann. 2017.
\newblock Continuous multilinguality with language vectors.
\newblock In \emph{EACL}.

\bibitem[{Pearl(2009)}]{causality}
Pearl, Judea. 2009.
\newblock \emph{Causality}.
\newblock {Cambridge University Press}.

\bibitem[{Ponti et~al.(2018)Ponti, O'Horan, Berzak, Vuli{\'c}, Reichart,
  Poibeau, Shutova, and Korhonen}]{ctsurvey}
Ponti, Edoardo~Maria, Helen O'Horan, Yevgeni Berzak, Ivan Vuli{\'c}, Roi
  Reichart, Thierry Poibeau, Ekaterina Shutova, and Anna Korhonen. 2018.
\newblock {Modeling Language Variation and Universals: A Survey on Typological
  Linguistics for Natural Language Processing}.
\newblock \emph{arXiv preprint arXiv:1807.00914}.

\bibitem[{Rabinovich, Ordan, and Wintner(2017)}]{haifa_transl:2017}
Rabinovich, Ella, Noam Ordan, and Shuly Wintner. 2017.
\newblock {Found in Translation: Reconstructing Phylogenetic Language Trees
  from Translations}.
\newblock In \emph{ACL}.

\bibitem[{Roberts and Winters(2013)}]{roberts:2013}
Roberts, Se{\'a}n and James Winters. 2013.
\newblock {Linguistic Diversity and Traffic Accidents: Lessons from Statistical
  Studies of Cultural Traits}.
\newblock \emph{{PloS One}}, 8(8):e70902.

\bibitem[{Serva and Petroni(2008)}]{serva:2008}
Serva, Maurizio and Filippo Petroni. 2008.
\newblock {Indo-European languages tree by Levenshtein distance}.
\newblock \emph{EPL}, 81(6):68005.

\bibitem[{Straka, Hajic, and Strakov{\'a}(2016)}]{udpipe}
Straka, Milan, Jan Hajic, and Jana Strakov{\'a}. 2016.
\newblock {UD-Pipe: Trainable pipeline for processing CoNLL-U files performing
  tokenization, morphological analysis, POS tagging and parsing}.
\newblock In \emph{LREC}.

\bibitem[{Velupillai(2012)}]{velupillai:2012}
Velupillai, Viveka. 2012.
\newblock \emph{{An Introduction to Linguistic Typology}}.
\newblock John Benjamins Publishing.

\bibitem[{Verma and Pearl(1992)}]{verma:1992}
Verma, Thomas and Judea Pearl. 1992.
\newblock An algorithm for deciding if a set of observed independencies has a
  causal explanation.
\newblock In \emph{Uncertainty in Artificial Intelligence}.

\bibitem[{Vinyals et~al.(2015)Vinyals, Kaiser, Koo, Petrov, Sutskever, and
  Hinton}]{Vinyals2015GFL}
Vinyals, Oriol, Lukasz Kaiser, Terry Koo, Slav Petrov, Ilya Sutskever, and
  Geoffrey Hinton. 2015.
\newblock {Grammar As a Foreign Language}.
\newblock In \emph{NIPS}.

\end{thebibliography}
